\documentclass[a4paper]{article}
\usepackage{graphicx}
\usepackage{mathtools}
\usepackage{amssymb}
\usepackage{authblk}
\usepackage{float}
\usepackage{times}

\title{The Concept of Forward-Forward Learning Applied to a Multi Output Perceptron }
\author[1]{K. F. Karlsson \thanks{Correspondence to: fredrik.karlsson3@saabgroup.com}}
\affil[1]{Saab Dynamics, SE-581 88 Linköping, Sweden}

\date{06 April 2023}

\begin{document}

\maketitle

\begin{abstract}
The concept of a recently proposed Forward-Forward learning algorithm for fully connected artificial neural networks is applied to a single multi output perceptron for classification. The parameters of the system are trained with respect to increased (decreased) "goodness" for correctly (incorrectly) labelled input samples. Basic numerical tests demonstrate that the trained perceptron effectively deals with data sets that have non-linear decision boundaries. Moreover, the overall performance is comparable to more complex neural networks with hidden layers. The benefit of the approach presented here is that it only involves a single matrix multiplication.

\end{abstract}

\section*{}

The perceptron is a type of artificial neural network that was introduced in the late 1950s by psychologist Frank Rosenblatt as a model of a biological neuron \cite{rosenblatt1958}. It is a single-layer feedforward neural network that learns from examples to classify input patterns by adjusting the weights of its connections. The perceptron has been widely used in pattern recognition and machine learning applications due to its simplicity and efficiency.

Despite its early success, the perceptron has a major limitation: it can only learn to distinguish linearly separable patterns \cite{minsky1969}. This means that it can only classify patterns that can be separated by a hyperplane in the input space. Patterns that are not linearly separable, such as the Boolean exclusive OR (XOR) function, cannot be learned by a single perceptron.

To overcome this limitation, multiple-layer neural networks were developed, such as the multilayer perceptron (MLP), which consists of several layers of interconnected neurons where the input to an interior hidden layer is given by the output of the preceding layer. These networks can learn non-linear decisions boundaries that separate the patterns, and it has been shown that nearly any function, to arbitrary accuracy, can be approximated by such a network with one hidden layer of sufficiently many neurons \cite{cybenko1989}.

Supervised training of neural networks is today typically accomplished by backpropagation, a learning algorithm that adjusts the  weights of the connections based on the error between the actual output and the desired output. Backpropagation works by propagating the error backwards through the network, layer by layer, allowing all its weights to be adjusted in a way that minimises the error \cite{rumelhart1986}.

For a typical multiclass classification problem, the input dimension of the perceptron is given by the dimension of the input signal, and the dimension of the output vector is equal to the number of classes. For a given input, the output value for each dimension can be interpretable by the softmax function as the probability for the input signal that it belongs to the corresponding class.
 
Recently, G. Hinton proposed the Forward-Forward (FF) algorithm that instead of minimising the final error at the network output optimizes the "goodness" of each involved network layer individually \cite{hinton2022ff}. Thus, this algorithm does not rely on backpropagation of information to preceding layers during learning. The goodness of a layer is a measure of the length of the output vector, and can be defined as the summed square of its elements. The aim is that \textit{good} or \textit{bad} input data should yield \textit{high} or \textit{low} goodness, respectively (referred to by Hinton as \textit{real} and \textit{negative} data). In this context, the procedure of supervised learning of a classifier is to provide the input not only with the training signal, but also with the the actual class label. If more than two classes are involved, one hot encoding is employed. As a consequence, the input vector is extended by one extra dimension per class. Input data with correct labels correspond to good data, and input data with incorrect labels correspond to bad data. 

In the above-mentioned publication, a network with four hidden layers of 2000 neurons each were reported to yield a performance comparable to state-of-art networks conventionally trained by backpropagation \cite{hinton2022ff}. Only the activity vectors of the last three hidden layers are combined in order to extract a collective goodness value, whereas the goodness of the first layer is omitted. By excluding the first layer, the overall performance reported to increase.

\begin{figure}[H]
\centering
\includegraphics[scale=0.20]{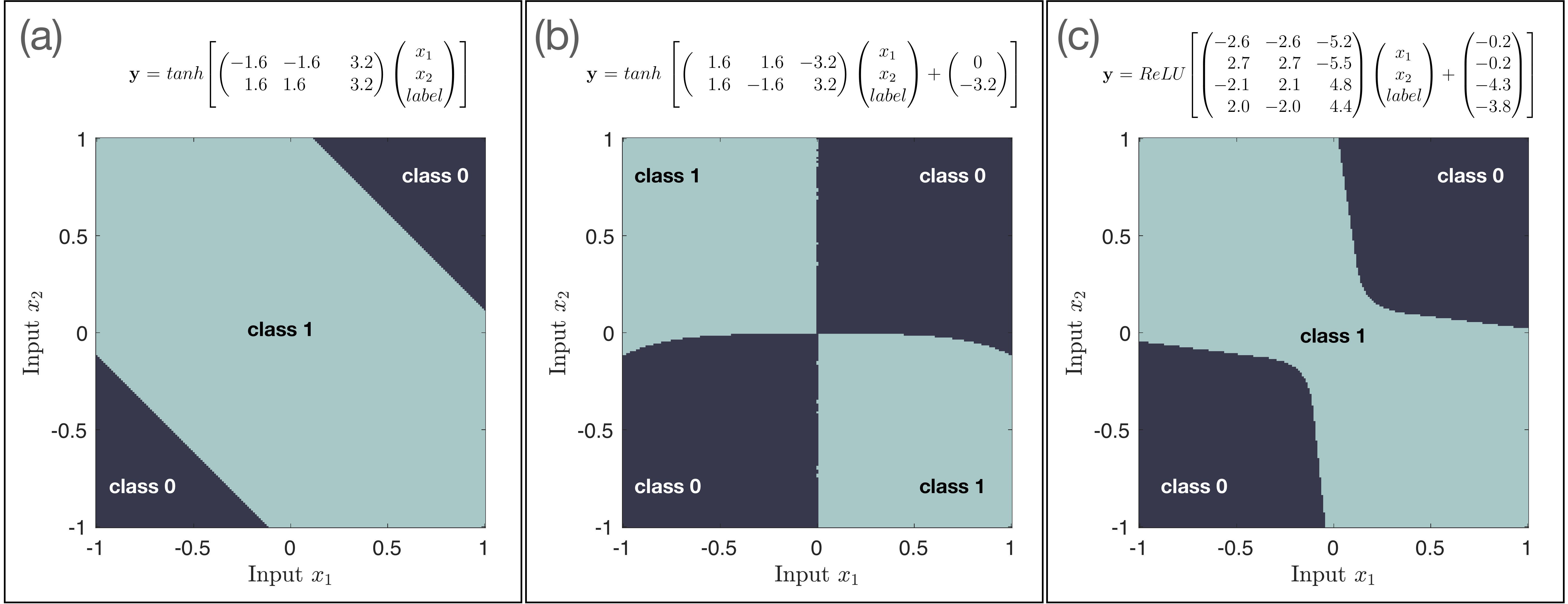} 
\caption{Classifications maps of a perceptron with two (a and b) or four (c) outputs trained to mimic the XOR function. Boolean true and false are represented by values near -1 or near +1 at the two inputs $x_{1}$ and $x_{2}$, respectively. The two classes are labelled by 0 for equal inputs, and 1 for different inputs. Given $x_{1}$ and $x_{2}$, the classification is based on the label that yields highest goodness, a value defined as the sum of the squared output vector elements $\mathbf{y}$. Above each map is the corresponding mathematical expression of the perceptron, including values for the trained parameters.}
\label{fig xor}
\end{figure}

In the present work, a single perceptron trained and used within the FF framework is studied. The training is performed with the Adam variant of stochastic gradient descent \cite{kingma2015}, to minimise the logistic loss applied to the goodness value minus a threshold. First, it will be shown that a single FF perceptron can be trained to mimic the XOR function. In this case, at least three inputs are required for training, the two XOR inputs and the class label. After training, the perceptron is expected to yield high goodness only if the two XOR input values correspond to the correct label, also given as input. Otherwise, low goodness is expected. Thus, this perceptron constitutes a binary classifier with the two categories of \textit{correctly} and \textit{incorrectly} labelled inputs. Since these two categories are not linearly separable in the three-dimensional input space, the expected goodness cannot be achieved by a single output. By using multiple outputs, however, the goodness value is determined by the collective properties of all outputs and this enables non-linear decision boundaries. As a consequence, a single FF trained perceptron, with more than one output, can perform XOR classification (see Fig. \ref{fig xor} for some examples with different number of outputs and activation functions). Thus, adding outputs to this perceptron is analogous to adding neurons to a hidden layer in a conventional MLP. 

\begin{figure}[H]
\centering
\includegraphics[scale=0.25]{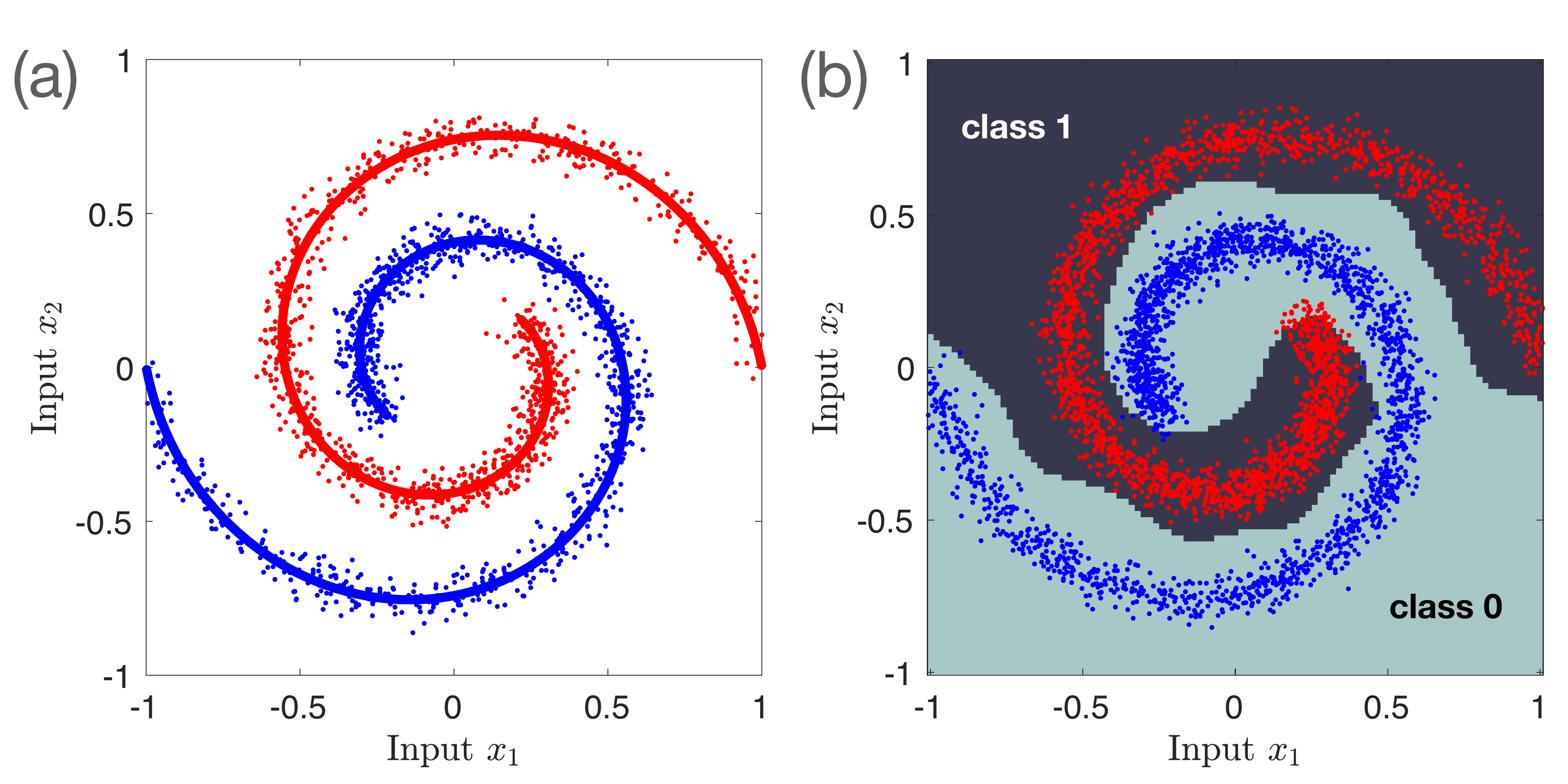} 
\caption{ Training data scattered around spiralling curves (a) of blue and red points belonging to two classes labelled 0 and 1, respectively, and (b) the resulting classification map of a trained perceptron with 3 inputs and 32 outputs. More than 99\% of random tests points are correctly classified by the trained perceptron.}
\label{fig2}
\end{figure}

A slightly more complex two-dimensional example is based on classifying noisy input points $(x_1, x_2)$ scattered along two blue and red spiralling curves according to Fig. \ref{fig2}a. Points of different colour belong to different classes. Besides $x_1$ and $x_2$, the class label should also be provided at the input. It is here given at a single input, label 0 for blue and label 1 for red, elevating the input dimension to three. With 32 leaky $ReLU$ activated outputs, the perceptron has 128 trainable parameters (including 32 biases). It classifies more than 99 \% of random test points correctly after training (see Fig. \ref{fig2}b). 
 
The previous examples clearly demonstrate that a single multi-output perceptron, trained and used within the framework of FF, is able to handle data sets with complex decision boundaries. Moreover, the number of parameters can be made arbitrary large by just increasing the number of outputs, and the performance is expected to improve with more parameters. As a final basic example, with input and output spaces of significantly higher dimensions, the perceptron with $ReLU$ activated outputs was trained to classify images of handwritten digits using the famous MNIST dataset \cite{deng2012mnist}. 60000 images were used for training and 10000 different images were used for evaluating the trained perceptron. The training set was split into mini-batches of size 10, and the training was performed such that 10 training image first was provided with correct labels (one-hot encoded), as good data, and subsequently the same images were provided with a random incorrect label as bad data. Gradients were computed and weights were updated after each mini-batch. See the code for more details \cite{github}.

Results obtained for some selected numbers of outputs are summarised in Table \ref{tab mnist}, with a test error below 1.7 \% for 2000 or more outputs. Augmenting the training data by randomly jittering the images up to two pixels in any direction, reduces the test error below 1.0\% for the perceptron with 8000 outputs. 

\begin{table}[b] 
\begin{tabular}{ |c|c|c|c| } 
 \hline
 Outputs & Data Augmentation & Test Error $\lesssim$ (\%) & Epochs\\
 \hline
 125 & none & 2.6 & 80\\ 
 500 & none & 1.9 & 80\\ 
 2000 & none & 1.7 & 150 \\
 8000 & none & 1.6 & 150\\
  \hline
 125 &  jittering & 2.2 & 180\\
 500 &  jittering & 1.4 & 260\\
 2000 & jittering & 1.1 & 340\\
 8000 & jittering & 1.0 & 340\\
 \hline
\end{tabular}
\caption{Results obtained on classification of the MNIST validation data after FF training of a single perceptron with various number of outputs. After the listed number of epochs, no significant improvement of the test error could be observed. A  learning rate of 0.0003 was used, the goodness was computed as the mean of the squared output elements and threshold in the logistic function was set to 10 \cite{github}.}
\label{tab mnist}
\end{table}

In many respects, the single FF perceptron studied here behaves as a conventional two-layer perceptron, i.e., with one hidden layer. It is considerably slower to train, however, and it seems to adapt less accurately to the training data than the conventional MLP. This property may make the FF perceptron less prone to overfitting.

In the future, it is expected that analogue neural networks will play an important role, mainly due to their predicted speed and low energy consumption. The single matrix multiplication involved in the evaluation of the perceptron outputs, combined with the absence of back propagation of errors between layers, makes the hardware architecture for analogue matrix computations much simpler. Indeed, single trainable perceptrons have already been realized in integrated memristor chips \cite{cai2019}, although still with very few inputs and outputs. Thus, high performing single perceptrons, as the one studied here, may find use for fast and energy efficient real time signal classification by future analogue hardware. 

Artificial neural networks are often regarded as black-box systems, which can be problematic in critical domains such as medical or military applications due to their lack of explainability. However, a perceptron that uses a single matrix for input-output mapping provides a transparent link between inputs and outputs, which promotes interpretability and explainability of the network. This transparency makes it easier to both visualize and understand the factors that influence outputs of the network.

In conclusion, it has been demonstrated that the Forward-Forward learning procedure enables a single perceptron to be trained to mimic the XOR function and to handle more complex and non-linear decision boundaries. For basic tests made on the famous MNIST data set of handwritten digits, the performance is comparable to conventional neural networks with hidden layers, and a classification test error smaller than 1\% was achieved. The simplicity of a single matrix multiplication promotes explainablity, and combined with the absence of backpropagation it opens up for simpler analogue computational hardware.

\section*{Acknowledgements}
The author is thankful to C. Larsson for bringing the recently proposed Forward-Forward algorithm to his attention, and to A. Gällström for interesting discussions on this topic.

\end{document}